\documentclass{article}
\usepackage[T1]{fontenc}
\usepackage[utf8]{inputenc}
\usepackage[]{ismir} % Remove the "submission" option for camera-ready version
\usepackage{amsmath,cite,url}
\usepackage{graphicx}
\usepackage{color}

\usepackage{subcaption}

\title{Gregorian melody, modality, and memory: \\ Segmenting chant with Bayesian nonparametrics}

\oneauthor
  {Vojtěch Lanz \hspace{1cm} Jan Hajič jr.}
  {Charles University, Faculty of Mathematics and Physics\\
   Institute of Formal and Applied Linguistics\\
   \texttt{\{lanz,hajicj\}@ufal.mff.cuni.cz}}
\def\authorname{V. Lanz and J. Hajič jr.}

\begin{document}

\maketitle

% \begin{itemize}
%     %\item % something
% \end{itemize}

\begin{abstract}
The idea that Gregorian melodies are constructed from some vocabulary of segments has long been a part of chant scholarship. This so-called ``centonisation'' theory has received much musicological criticism, but frequent re-use of certain melodic segments has been observed in chant melodies, and the intractable number of possible segmentations allowed the option that some undiscovered segmentation exists that will yet prove the value of centonisation, and recent empirical results have shown that segmentations can outperform music-theoretical features in mode classification. 
Inspired by the fact that Gregorian chant was memorised, we search for an optimal unsupervised segmentation of chant melody using nested hierarchical Pitman-Yor language models. 
The segmentation we find achieves state-of-the-art performance in mode classification. 
Modeling a monk memorising the melodies from one liturgical manuscript, we then find empirical evidence for the link between mode classification and memory efficiency, and observe more formulaic areas at the beginnings and ends of melodies corresponding to the practical role of modality in performance. However, the resulting segmentations themselves indicate that even such a memory-optimal segmentation is not what is understood as centonisation.
% there are limits
% to how informative any selection of melodic segments can be with respect to mode:
% modality might not be solely determined by the melody itself.
%
% Contributions:
% Memorisation-based unsupervised segmentation of chant melody.
% setting the state-of-the-art for segmentation-based mode classification
% % JH: "through more careful consideration of Gregorian chant practices..."
% % This says we found a useful segmentation!
% NHPYLMClasses extending the Hierarchical Pitman-Yor model.
% %
% % JH:
%
% % JH: into discussion. ...precisely where antiphons and responsories interface with psalm tones and responsorial verses.
%
\end{abstract}

\section{Introduction}
\label{sec:introduction}
Gregorian chant, the liturgical monody of the Latin church, has been a pillar
of European musical identity since the Carolingian period in the 8th-9th century A.D. 
%However, it is still eluding efforts at a systematic description of its musical properties: 
And yet we have no constructive theory of chant melody 
like we have e.g., tonality for classical composition.
%
% We do know chant melody is tightly coupled to its text, in the sense that text-melody
% pairs are prescribed by the Roman rite (and other rites) for individual events in liturgy,
% such as the \textit{introit} at the beginning of mass, 
% even though the exact nature of the relationship between melody and text is still being elucidated \cite{cornelissen2020mode,lanz2023}.
%
% We know that different classes of liturgical occasions tend to attract melodies with a similar
% level of ornamentation, measurable primarily through how melismatic a melody is -- the proportion
% of the length of a melody in pitches to the number of syllables in its associated text.
%
No good explanation exists for why Gregorian melodies \textit{should} be the way they are,
or how to write a plausible one from scratch.
``They are not meant to be composed at all'' would be an answer,
%given the legend of their divine origin \cite[p.511]{hiley1993western} --
had no Gregorian melodies been newly composed or heavily edited. 
However, new chants were composed \cite[p.~463]{hiley1993western},
%, going well beyond simple contrafacts,\footnote{Contrafacts re-use a melody but set a different text to it.} 
for example, for new feasts \cite{hallas2021offices}.
Thus, the question remains: how is Gregorian melody structured?

\subsection{Modality}

% Modes as in music theory
%The closest we have to a theory of Gregorian melody is modality. %, 
%contemporaneous with its Carolingian roots and present throughout its development. 
%Within the field of music information retrieval, 
%Modality is most straightforwardly formalised as a category to which melodies belong.
%There are eight basic modes, 1--8,\footnote{Leaving aside transposed modes etc.}
%also known as the church modes and 
%today most often referenced by their Greek names: 
%dorian, hypodorian, phrygian, etc. %,
The main theoretical framework for Gregorian melody is modality, which classifies melodies into eight basic modes 1--8,\footnote{Leaving aside transposed modes etc.} also known as the church modes and today most often referenced by their Greek names: dorian, hypodorian, phrygian, etc.
%\footnote{In medieval sources, modes are referred to rather by number: \textit{protus}, \textit{deuterus}, \textit{tritus}, and \textit{tetrardus}, with the variants \textit{authethus} and \textit{plagalis}. Additionally, the \textit{tonus perergrinus} exists, but that is mostly specific to psalm tones. Furthermore, in the course of the middle ages a mix of evolving music theory and scribal necessity led to the introduction of transposed modes \cite{hiley1993western}.}
%
%Especially since the 11th-century \textit{Micrologus} of Guido of Arezzo \cite{merkley1988families},
Modes are primarily identified by their \textit{finalis} (the final note, typically $d$, $e$, $f$, or $g$) and their range (\textit{authentic} or \textit{plagal}). 
%Melodies in modes 1 and 2 end on $d$, 3 and 4 on $e$, 5 and 6 on $f$, and 7 and 8 on $g$. The range distinguishes authentic (odd) and plagal (even) modes: roughly a 7th or octave above the finalis for authentic modes, and a 4th below and a 5th above for plagal ones. 
This ``theoretical'' definition 
is passed down from early medieval sources throughout the Middle Ages \cite[Ch.~1]{merkley1988families,atkinson2008critical}. 

%Thusly defined, modality is as a principle of organising \textit{melodies}.
The medieval definition of modes % is unsatisfactory as a theory of melody: it
says little about how Gregorian melody should be constructed.
Modality involved more than just the beginnings, ends, and ranges of melodies: an evolving understanding of modality \cite[ch.~6]{atkinson2008critical} led to revisions, most notably, the Cistercian order made extensive revisions based on their notion of modal ``purity'' \cite[p.~610]{hiley1993western} \cite[p.~72]{glasenapp2020pray}.
Tonaries\footnote{Manuscripts that explicitly organise melodies by mode.} \cite{huglo1971tonaires,merkley1988families} show that the repertoire often diverged from the ``theoretical'' definition.
%Looking at contemporaneous evidence from tonaries\footnote{Manuscripts that explicitly organise melodies by mode.} \cite{huglo1971tonaires,merkley1988families}, 
%the Gregorian repertoire does not entirely conform to the %``theoretical'' definition of mode.
At least 1 in 10 antiphons and responsories would be misclassified using the ``theoretical'' features \cite{cornelissen2020mode}. Some tonaries even disagree with each other \cite{mori2001conflicting}.  Mode, therefore, goes beyond its medieval definition.
%and practical modality has been seen as a problem.

An attractive approach to modality is the idea that 
%beyond the \textit{finalis}, range, possibly initial,
%and a preference for using some tones over others, 
modes are hidden ``vocabularies'' of characteristic melodic segments.
This follows from observations that many melodic gestures are re-used
across different melodies \cite{helsen2009use,karp1998aspects,Helsen2023}.
Melodies within a mode tend to be similar \cite{gevaert1895melopee,merkley1988families},
suggesting these segments may be specific to modes.
Text-based and other naive segmentations of chant melodies have been used for mode classification \cite{cornelissen2020mode,lanz2023}, showing promising results on the CantusCorpus v0.2 dataset \cite{lacoste2022cantus, cornelissen2020studying} and outperforming ``theoretical'' approaches, as well as pitch profiles.
Taken to the extreme, the theory of ``centonisation'' postulates entire melodies are constructed by concatenating mode-specific melodic segments \cite{ferretti1934estetica,levy1970italian}.

Centonisation has faced strong criticism in Gregorian chant scholarship \cite[p.~74--75]{treitler1975centonate,hiley1993western}; however, this largely concerns its framing as a deliberate compositional strategy, not the recurrence of melodic material across chants \cite{treitler1975centonate}.

\subsection{Memory}

Gregorian chant was originally an oral tradition  \cite{treitler1975centonate, hucke1980toward, atkinson2008critical}, requiring monks to memorise thousands of melodies without exact pitch notation for over 300 years. The challenge of memorisation led to the rapid adoption of staff notation, introduced by Guido of Arezzo in the 11th century as a teaching aid  \cite[p.~388]{atkinson2008critical,hiley1993western}.

%In this paper, we propose a method for finding one such ``optimal'' segmentation
%based on the performance practice of chant: 
%Recall that Gregorian chant was, % a memorised tradition,
%for much of its medieval existence, 
%especially during its early formative centuries \cite{hucke1980toward}, an oral tradition, requiring monks to memorize chants \cite{treitler1975centonate,atkinson2008critical}. 
%This is most obviously evidenced by the absence of exact
%pitch notation for the first 300+ years of its existence.
%The repertoire of a liturgical year involved several thousand melodies, making memorization a challenge. This importance is evidenced by the introduction of staff (diastematic) notation by Guido of Arezzo in the mid-11th century. Originally as a pedagogical tool to enable learning melodies without a teacher, this notation system was adopted very quickly \cite[p.~388]{atkinson2008critical,hiley1993western}.

% Centonisation over time
%In this situation
Memory constraints, along with cultural-evolutionary and information-theoretic principles, strongly support the emergence of centonisation in the melodies.
Oral transmission is imperfect \cite{shanahan2019examining,anglada2022studying}, and melodies tend to evolve toward lower entropy \cite{popescu2024evolution}.
Since melodies were rarely written the same way twice \cite{froger1978critical,hajic2023}, despite efforts to preserve them \cite[p.~611]{hiley1993western}, they changed during centuries of oral transmission before getting written down with exact pitch notation. Given the memorisation demands on singers, centonisation would be expected under the Minimum Description Length (MDL) principle. Supporting this, a recent study~\cite{anglada2022studying} found that oral transmission led to \textit{``using fewer building blocks (intervals, contours) that are increasingly reused and combined''} over time, which is exactly the type of structure centonisation refers to.

%Many other monodic traditions with oral transmission do exhibit centonized repertoire.
%Centonisation has been observed in
Many monodic oral traditions show centonisation -- for example, Arab-Andalusian \cite{nuttall2019contributing} and Byzantine chant \cite{wellesz1947words}, and high formulaicity has been observed in Old Roman \cite{connolly1972introits,dyer1998tropis} and Beneventan chant \cite{huglo1985beneventan}. 
Melodic formulas appear in Hindustani Ragas \cite{chakraborty2021hindustani},
and we are certainly omitting tens, if not hundreds, of other such traditions here.

% While the results of \cite{cornelissen2020mode} that textual boundaries 
% specifically lead to segments that classify Gregorian mode better than others does not in fact hold
% for antiphons \cite{lanz2023}, segmentations do clearly outperform the ``traditional''
% music-theoretical features in mode classification at least on responsories.

\subsection{Outline}
Since testing all segmentations is intractable, it remains possible that \textit{some} segmentation could reveal that Gregorian chant is a ``centonate'' tradition.

%Given that testing all segmentations of the repertoire is intractable, 
%an enticing possibility still exists that \textit{some} segmentation of chant melodies exists 
%that would in fact show that Gregorian chant is a ``centonate'' tradition, like others.

In Section~\ref{sec:methods}, we formalise chant memorisation as a computational segmentation problem, searching for an ``optimally centonised'' segmentation of chant by exploiting the connection between the MDL principle \cite{rissanen1998stochastic,grunwald2007minimum} 
and Maximum a Posteriori (MAP) estimation \cite{barron1991minimum,Goldwater2009Bayesian},
and ``borrowing'' the nonparametric Bayesian Nested Hierarchical Pitman-Yor Language Model (NHPYLM)
\cite{teh2006} to segment chant melody instead of natural language.

After briefly introducing the datasets we use (Section~\ref{sec:data}),
using mode classification as a proxy for segmentation quality \cite{cornelissen2020mode,lanz2023},
we show that the memorisation-based method outperforms existing segmentation baselines,
both on pitch and interval representations (Section~\ref{sec:experiments}).

Imitating a monk learning from one manuscript (Section~\ref{sec:dkaug}), we find that efficient memorisation relates to modality, though its influence varies across melody parts, depending on how they were performed in liturgy
%Imitating a monk learning melodies from one manuscript %(Section~\ref{sec:dkaug}),
%we find empirical indications that efficient memorisation relates to modality,
%but modality does not necessarily influence all parts of the melody equally, depending on how they were performed in liturgy.
%Notably, the longer and more virtuosic responsories turn out to be easier to remember than the shorter antiphons.

We also contribute a Cython implementation of the NHPYLM model, and a class-conditioned version thereof.\footnote{Available at \url{https://github.com/lanzv/nhpylm}}

\section{NHPYLM--Related Work}
\label{sec:relatedwork}

In tasks such as vocabulary induction, phoneme-to-word mapping, and word segmentation of text without whitespace (e.g., Mandarin or Japanese), nonparametric Bayesian methods avoid manual model selection by learning model complexity from the data as part of inference.
%In tasks like chant melody segmentation, where the number of segments is unknown and there is not enough data for deep learning-based approaches, Bayesian nonparametric statistical methods have shown .  especially as it is low-data scenarios. 
In this context, the Hierarchical Pitman-Yor Language Model (HPYLM) \cite{teh2006} was introduced. %as an extension of the Dirichlet Process. % more effectively modelling natural language sequences. 
It has been widely applied to unsupervised word segmentation, from speech recognition \cite{walter2013unsupervised,takeda2018word} to topic models \cite{araki2012online}.
%capturing longer-range dependencies by recursively structuring probability distributions. 
Mochihashi et al. \cite{mochihashi2009} extended this approach with the Nested HPYLM (NHPYLM), incorporating character-level priors from Character HPYLM.

NHPYLM has been adapted for melody segmentation, modelling motifs in tonal music \cite{sawada2020unsupervised}. The segmentation results were compared with the Generative Theory of Tonal Music, demonstrating the effectiveness of Bayesian nonparametric models for structured sequence learning, which we extend in the context of Gregorian chant.

% Modality in chant has been...

\section{Memory-efficient segmentation}
\label{sec:methods}

% The Minimum Description Length (MDL) principle, in its basic two-part form \cite{},
% says the optimal model in terms of compressing some dataset $X$ can be found
% by minimising $L(D|M) + L(M)$, where $L(D|M)$ is the length of the data encoded by
% some model $M$ and $L(M)$ is the length of the encoded model.
% $L(D|M)$ is, via Shannon's theorem, minimised by defining $M$ to be a probability distribution
% from a given family $f$ with parameters $\hat{\theta}$ found via Maximum Likelihood estimation on observed data $D$ \cite[p.~11]{hansen2001MDL}.

Efficient memorisation of the Gregorian melodies can be formalised with the MDL principle \cite{rissanen1998stochastic,grunwald2007minimum}: the best code $H$ for data $D$ is one that minimises $L(D|H) + L(H)$, where $L(D|H)$ is the length of the data encoded using the codebook $H$, and $L(H)$ is the length of the codebook. The MDL principle thus encourages codebooks such that the more frequently occurring and the longer a subsequence, the shorter code it gets to minimise $L(D|H)$, and it encourages small codebooks via the $L(H)$ term. % ensures that the degenerate case where every sequence is segmented as itself has a codebook the size of the original data, so no memory efficiency is achieved.
Importantly for inferring memory-efficent segmentations, choosing a hypothesis using the MDL principle is equivalent to choosing the MAP hypothesis under a Bayesian model \cite{barron1991minimum}, where the prior probability of a hypothesis (corresponding to $L(H)$) decreases 
%exponentially 
with its length \cite{Goldwater2009Bayesian}.

%For the task of unsupervised segmentation model, the Pitman-Yor process \cite{PitmanYor1997} fulfils this condition, and its nested hierarchical version can model sequential dependencies of inferred segments, in an analogy to n-gram language models \cite{teh2006}.

%Finding an optimal model under the Minimum Description Length principle is equivalent to MAP estimation \cite{barron1991minimum}, with the length of the codebook (model) following from the prior probability of the model.
To find an optimally memory-efficient segmentation, 
we need an unsupervised segmentation model that: 
(1) infers the vocabulary, including its size, since we want the vocabulary of chant
segments to be a dependent variable following from efficient memorisation; 
(2) can model the sequential nature of chant melody (as one sings one segment after the other); 
(3) has a prior that prefers smaller segment vocabularies; 
and (4) has tractable inference and MAP estimation.
These conditions are fulfilled by the Nested Hierarchical Pitman-Yor Language Model (NHPYLM) \cite{teh2006}.

\subsection{An intuition on the NHPYLM}

The Pitman-Yor process (PY) is a nonparametric Bayesian model that assigns probabilities
to categorical distributions without setting the number of categories in advance \cite{PitmanYor1997}.
The intuition behind PY starts with the Chinese Restaurant Process (CRP):\footnote{Also: Dirichlet Process.}
the categorical distributions are normalised customer counts at tables serving a different dish each.
The key property of the CRP is that the $n$-th customer chooses to sit at a given table 
proportionally to how many customers are already eating there, 
with a parameter $\alpha$ that leaves a (decreasing) chance $\alpha / (\alpha + n - 1)$ of sitting at a new table.
This gives rise to a rich-get-richer behaviour, and a prior that strongly prefers fewer tables -- in the case of language models, fewer vocabulary elements and thus shorter hypotheses under the MDL principle.
PY is a generalisation of CRP that provides more control 
over the rich-get-richer behaviour, to better model long-tail distributions
such as those found in natural languages \cite{PitmanYor1997}.\footnote{PY differs from CRP by applying a constant discount to table counts, slowing the decay of the ``density budget'' for new tables.}

In sequence segmentation, PY finds the optimal vocabulary 
under this rich-get-richer paradigm over unigram probabilities of the resulting segments.
In order to model the sequence ordering with bigrams $P(s_i | s_{i-1})$ of the inferred segments, 
we must add a ``restaurant of restaurants'', which models the probability of having the history $s_{i-1}$. The ``inner'' restaurant then models the conditional probability of $s_i$,
via a yet another hierarchy of restaurants over the character n-grams in $s_i$.
This is, finally, the Nested Hierarchical Pitman-Yor Process \cite{teh2006,mochihashi2009}.

In our case, the ``inner'' HPYLM operates on tones, while the ``outer'' operates on segments.
The base distribution for the segment-level HPYLM is obtained from the tone-level HPYLM. 

\subsection{Segmentation Probability with NHPYLM}

To compute the probability of a given segmentation of a melody, we only need to evaluate the probability of each segment in its context under the NHPYLM and multiply them together. For a segment $s$ given a context $h$, the ``outer'' (segment-level) model computes the probability recursively as:

\begin{equation}
\label{eqn:segprob}
p(s|h) = \frac{c(s|h) - d \cdot t_{hs}}{\theta + c(h)} + \frac{\theta + d \cdot t_{h}}{\theta + c(h)} \cdot p(s|h'),
\end{equation}

\noindent
where $c(s|h)$ is the number of customers (segment occurrences) for segment $s$ with context $h$, $c(h)$ is the total number of customers with that context, $t_{hs}$ is the number of tables serving segment $s$ in context $h$, and $t_h$ is the total number of tables with context $h$. The discount $d$ and concentration $\theta$ are hyperparameters of the Pitman-Yor process. The term $p(s|h')$ refers to the probability of segment $s$ with a shorter context $h'$. We use a bigram model, so the initial context $h$ always consists of a single preceding segment, and $h'$ is the empty context.

When no shorter context remains, the base distribution is the ``inner'' tone-level model, which computes the probability of a segment composed of tones $t_1, \dots, t_k$ as:

\begin{equation}
\label{eqn:innerprob}
p(t_1 \dots t_k) = \frac{\prod_{i=1}^{k} p(t_i \mid t_1 \dots t_{i-1})}{p(k)} \cdot Po(k \mid \lambda),
\end{equation}

\noindent
where $p(t_i \mid t_1 \dots t_{i-1})$ is the tone-level probability computed by marginalizing over all possible context lengths $n$:

\begin{equation}
\label{eqn:tonemarginal}
p(t \mid h) = \sum_{n=0}^{\infty} p(t \mid h, n) \cdot p(n \mid h),
\end{equation}

\noindent
Here, $p(t \mid h, n)$ is the probability of tone $t$ given tone context $h$ at depth $n$ (i.e., the last $n$ tones of the context are considered). This is computed using Equation~\ref{eqn:segprob}, but applied at the tone level with a uniform base distribution over tones. The term $p(n \mid h)$ denotes the probability that context $h$ has order $n$.
$Po(k \mid \lambda)$ is the Poisson correction to prevent short segments from being over-favored \cite{mochihashi2009}.
Further mathematical details can be found in the original formulations of the model \cite{teh2006, mochihashi2009}.

The model is trained using blocked Gibbs sampling. Initially, each melody is assigned a random segmentation.
%These initial segmentations are added to the NHPYLM model, forming a probability distribution. 
In each sampling step, a randomly selected segmented melody is removed from the model,
all its possible segmentations are evaluated with eqs. (1)--(3) using the current state of the model,
and a new segmentation is sampled from them.
%and re-segmented according to the current probability distribution of NHPYLM. 
%, and resampled proportionally to the NHPYLM distribution. 
The model is then updated with the newly sampled segmentation.
During inference, optimal segmentation is computed using the Viterbi algorithm.

\subsection{Mode-specific segmentation}
\label{subsec:nhpylmclasses}

%Given how ``small'' tonaries organise memorisation of Gregorian repertoire by mode \cite[ch.~3]{huglo1971tonaires,merkley1988families}, 
%we also experiment with training a separate NHPYLM over each mode's melodies.
Each mode may have its own characteristic melodic units and segmentation patterns
\cite{ferretti1934estetica,levy1970italian}.
%\textbf{TODO:Evidenced by tonaries.}
%
We extend the NHPYLM model to account for this by training separate segmentation models for each mode, as though one were learning the repertoire in 8 separate sub-corpora.
%We create eight NHPYLM models, each trained on chants of only a single mode. 
% During training, a model knows the chant's mode and uses it to guide segmentation. 

At inference time (i.e., when analyzing new chants), the mode of a chant $c$ is unknown. However, we can use all eight mode-specific models (each estimating $p(\bar{c} \mid m)$, the likelihood of the optimal segmentation $\bar{c}$ under mode $m$) to determine the most probable mode $m^* = \arg\max_m p(m \mid c)$ given the melody $c$, using Bayes' Theorem.\footnote{Mathematical and implementation details are available at: \\\url{https://github.com/lanzv/chant-modality-with-nhpylms}
%The prior on $p(m)$ is set to uniform.
} 
% Conditioning mode on the melody itself requires marginalising over all segmentations $\bar{c}$, which is intractable; we approximate it using only 8 segmentations $\bar{c}^{*}_{l}$ that are the optimal segmentations of $c$ under each of the 8 modes $l = 1 \dots 8$. Once we have the estimated $m^{*} = argmax_m p(m|c)$, we find the optimal segmentation $\bar{c}^{*}$ as a pointwise estimate using the NHPYLM for mode $m^{*}$ -- which we have in fact already computed when approximating $p(m|c)$ for $l = m^{*}$. 
We refer to the combined model as \textit{NHPYLMClasses}.

\subsection{From Melody Segments to Modes}

Given the interaction between mode, melodic similarity, and memorisation \cite{lanz2023}, mode classification based on inferred segments can serve as a proxy for segmentation quality. However, aside from the NHPYLMClasses model, this ``amount of information retained about mode'' must be obtained in a separate downstream step.
% training a mode classifier over the segmentations.
%
%The previous Section~\ref{subsec:nhpylmclasses} described one possible approach for classifying modes based on segmentation. However, such classification is not available in the case of the basic NHPYLM model described in Section~\ref{subsec:nhpylm} or from any of the segmentation baselines such as words, etc. \cite{cornelissen2020mode}. 

To ensure comparability with previous work, we adopt the same mode classification pipeline used in earlier studies of chant segmentation and modality \cite{cornelissen2020mode,lanz2023}. Each melody is represented as a bag-of-segments vector. Although sequential information is not explicitly encoded, segments are inferred using a model that considers sequence context, so some information about order is implicitly preserved. TF-IDF weighting is applied to the vectors.
%to prevent the classification from reducing to a simple ``tone profile'' approach, even when short segments (e.g., single pitches) are used.
% term-frequency inverse document-frequency (tf–idf) vectorizer, a standard weighting scheme in information retrieval. The tf–idf score for a given segment $s$ in chant $c$ is computed as follows:  
% \begin{equation}
% \text{tf–idf}(s, c) = \text{tf}(s, c) \cdot \log \frac{N}{\text{df}(s) + 1}
% \end{equation}
% where $\text{tf}(s, c)$ represents the frequency of segment $s$ in chant $c$, $N$ is the total number of chants, and $\text{df}(s)$ is the number of chants containing segment $s$. The addition of $1$ in the denominator ensures smooth inverse document frequency (idf) computation.  
%

Following settings used in previous work on chant segmentation \cite{cornelissen2020mode}, only the 5000 most frequent segments are retained, with others discarded. The resulting vectorized segmentations from the training melodies are then used to train a Linear SVM classifier to predict chant modality.

%This model assigns a mode to each segmented melody in the test set based on its vectorized segmentation. The classification performance is then evaluated using the F1-score by comparing the predicted and gold modes.

\section{Datasets}
\label{sec:data}

We use the \textbf{CantusCorpus v0.2} dataset \cite{cornelissen2020studying}, derived from the Cantus Database \cite{lacoste2022cantus}, the most extensive digital collection of Gregorian chants. We focus on the two most abundant genres: antiphons and responsories. Antiphons are simpler and shorter melodies, while responsories are more complex and longer.
%
%This database contains extensive information on thousands of medieval manuscripts, including metadata on melodies, genres, and modes, which are particularly relevant for this research. The melodies are stored in Volpiano format, and additional details such as final tones, differentiae, and lyrics are also available. Additionally, chants are mapped to their source manuscripts, specific pages, and liturgical contexts.
%
We apply the same preprocessing steps as used in previous work \cite{cornelissen2020mode}, keeping only completely transcribed melodies with simple mode annotations (1–8), and discarding non-pitch characters from Volpiano encoding.
Crucially, we also remove \textit{differentiae} from antiphons \cite{lanz2023}.
%
% However, not all melodies are fully transcribed, making it necessary to apply a filtration process to the CantusCorpus dataset \cite{cornelissen2020mode}. This process involves removing chants that lack Volpiano melodies, simple mode annotations, or a sufficient number of notes, as well as those with incomplete lyrics or an incipit used as the full text. Additionally, melodies that begin with a clef other than G, contain incorrect Volpiano characters, missing pitches, or improper word boundaries are discarded. 
% %
% Crucially, \textit{differentiae} were removed from antiphon melodies \cite{lanz2023}, and duplicate melodies were eliminated.
%
After the described filtration process \cite{cornelissen2020mode,lanz2023}, a total of $13551$ antiphons and $7031$ responsories remain. Because responsories are longer (137.38 pitches on average vs. 53.97 for antiphons), the responsory dataset is longer, at 966k notes, with antiphons at 731k notes.
%However, each of these melodies originates from $640$ sources, which also influence the structure of the melodies. 

The CantusCorpus also contains multiple versions of the same melody
if it is found in multiple transcribed sources. Although these are almost never identical \cite{froger1978critical,hajic2023}, this still means closely related melodies could end up
in both the test and training sets. While we could ensure this does not happen
using the CantusID mechanism \cite{lacoste2022cantus}, the abundances of different melodies
would still distort results if present in the test set, and distort optimisation if present
in the training set.
We propose using melodies from a single source. 
This resembles the situation of any chant practitioner: 
they would be expected to learn their local repertoire.
We chose the largest available manuscript, \textbf{D-KA Aug. LX},\footnote{\url{https://cantusdatabase.org/source/123612}}, with $1965$ antiphons and $907$ responsories. 

%Detailed statistics on the distribution of melodies across modes in both datasets are shown in Table~\ref{tab:dataset}.

\begin{table}[t]
    \centering
    \setlength{\tabcolsep}{5.5pt}
    \begin{tabular}{lcc|cc}
        \textbf{CantusCorpus v0.2 data} & \multicolumn{2}{c|}{\textbf{Pitches}} & \multicolumn{2}{c}{\textbf{Intervals}} \\ 
        \cline{2-5}
        \textbf{} & \textbf{Ant.} & \textbf{Res.} & \textbf{Ant.} & \textbf{Res.} \\
        \hline
        \textit{Classical approach} & \textit{89.6} & \textit{89.4} & -- & -- \\
        \hline
        4-gram & 91.0 & 91.6 & 82.0 & 83.1 \\
        Syllables & 89.3 & \textbf{\textit{93.5}} & 72.9 & 89.3 \\
        Words & 90.1 & 90.2 & 83.9 & 87.2 \\
        NHPYLM & 91.7 & 93.3 & 86.7 & 89.7 \\
        NHPYLM-Cl, no SVM & \textbf{\textit{92.6}} & \textbf{\textit{93.8}} & \textbf{90.4} & \textbf{92.4} \\
        NHPYLM-Cl & \textbf{92.7} & \textbf{93.9} & \textbf{\textit{90.2}} & \textbf{92.4} \\
        \hline
        \textit{Overlapping n-grams, 1-7} & \textit{93.8} & \textit{94.8} & \textit{90.4} & \textit{92.8} \\
    \end{tabular}
    \caption{F1 scores for various mode classification methods applied to all antiphon and responsory melodies, encoded as both sequences of pitches and sequences of intervals. %\textbf{TODO:finish description -- abbrevs etc.} 
    St.dev. across the 5 splits was between 0.001 and 0.005, so differences of an F1-score of 1.0 is ``safe'' at $1.96\sigma$.}
    \label{tab:all_f1}
\end{table}

\section{Mode classification experiments}
\label{sec:experiments}

In this section, we describe experiments on the CantusCorpus dataset.
%
%\textbf{Experimental setup.}
%Different segmentation approaches applied to melodies are compared and examined how well an SVM classifier can correctly assign these segmentations to modes. For baseline comparisons, we use simple segmentation methods, including word-based (\textit{Words}), syllable-based (\textit{Syllables}), and 4-gram-based (\textit{4-gram}) segmentations \cite{cornelissen2020mode}, excluding neume-based and other n-gram segmentations due to their weaker performance. 
%
%We compare these methods with segmentations generated by the NHPYLM and NHPYLMClasses methods, as described in Section~\ref{sec:methods}, which are used as input to the SVM. 
Both NHPYLM-based methods are configured with segment lengths ranging from 1 to 7 tones using default hyperparameters: $d_0=0.5$; max. segment length of 7; initial $\theta = 2.0$  for inner hyperparameter updates \cite{mochihashi2009}; Gamma priors on both $d$  and $\theta$ with $\alpha = 1.0, \beta = 1.0$; and a Gamma prior for the Poison correction parameter with initial $\alpha_{\lambda} = 6.0, \beta_{\lambda} = 1.2$.\footnote{See also Supplementary materials -- NHPYLM implementation.}
%
%The newly sampled segmentation is then added back to the model, which updates its probability distribution accordingly.
%
$10 \%$ of the training data is reserved as a validation set for %hyperparameter tuning and 
checking for convergence for NHPYLM-based models.

%As described in Section \ref{subsec:nhpylmclasses}, the NHPYLMClasses method additionally produces its own modal predictions, which we refer to as \textit{NHPYLMClasses, no SVM}.

%\textbf{Melody representations.}
Because relative pitch has greater saliency than absolute pitch \cite[p.56,p.100]{Hallam2017musicPsychology}, in addition to pitch sequences from cleaned CantusCorpus melodies, 
we also use sequences of intervals (which thus have a length of $n-1$).

For all experiments, we use a 7:3 train-test split for both NHPYLM-based models and the SVM classifier.
%(that is, even though NHPYLM models are unsupervised, we do not include the test set for mode classification into training).
In the case of NHPYLM and NHPYLMClasses models,
training data is used to learn the optimal probability distribution, 
and the resulting segmented training data is then used to train the SVM classifier.
(This is why the train-test split is necessary also for the unsupervised methods.)
During evaluation, the test set is first segmented by the NHPYLM models, after which the SVM classifier predicts the modes. 

\subsection{Evaluation}
\label{subsec:evaluation}

We evaluate how well the proposed segmentation methods retain modal information by using inferred segments as features in a mode classification task. This approach is musicologically justified, given the relationship between modality and memory \cite[ch.~3]{lanz2023,atkinson2008critical}, and has been used before \cite{cornelissen2020mode}. We report micro-averaged F1-scores in Tables~\ref{tab:all_f1} and~\ref{tab:dka_aug_f1}.
For NHPYLMClasses, we also report the model's ``internal'' classification performance, i.e., accuracy based on mode $m^* = \arg\max_m p(m \mid c)$ selected during inference, alongside SVM results using the inferred segments.

%Although segmentation methods have shown potential for effectively describing chant melodies \cite{cornelissen2020mode}, 
%it remains unclear what \textit{optimal} segmentation of Gregorian chant melodies should look like. 
\textbf{Baselines.}
Previous work \cite{cornelissen2020mode} segmented chants using n-grams, neumes, syllables, or words. 
We adopt their best-performing segmentations as baselines: 4-grams, syllables, and words. 
We also include the \textit{Classical approach} as a baseline, which classifies modes based on initial and final tones along with the melody range \cite{cornelissen2020mode}.

%Since no true segmentation is known, we cannot be sure whether crucial segments were omitted in our NHPYLM-based approaches. 
%
\textbf{Upper bound.}
We do not know %, given the uncertainties connected to the relationship between melody and mode, 
what the maximum achievable mode-classification performance over any segmentation is.
The mode to which a melody belongs may not be solely determined by the melody itself:
some melodies are assigned to different modes in different tonaries \cite{mori2001conflicting}.
We at least make a rough estimate of the upper bound of mode classification performance 
that can be achieved using a ``distributional approach'' \cite{cornelissen2020mode} 
and the SVM classifier as a reasonable ``information extraction'' black box.
In this setting, we use all possible overlapping n-grams of lengths 1 to 7 as features. Thus, more information is available to the classifier than for any segmentation (with TF-IDF preprocessing to limit the effect of many uninformative features). We refer to this approach as \textit{Overlapping n-grams, 1-7}.
Already overlapping 4-grams outperformed all previous results \cite{lanz2023}.

\textbf{Cross-validation.} Each experiment was repeated five times with different random seeds for a 70-30 random split over the entire dataset. While the split is not explicitly stratified by mode, averaging results over multiple random splits helps mitigate potential imbalances and ensures more robust performance estimates.

\subsection{Results}
\label{subsec:results}

Mode classification for CantusCorpus antiphons and responsories is shown in Table~\ref{tab:all_f1}. Two representations are used: sequences of pitches and sequences of intervals.

%\textbf{TODO:What differences are significant? We don't have confidence intervals...}

Across all four settings, NHPYLMClasses outperformed all baselines, 
though the syllable baseline on responsories with absolute pitches comes close. 
Notably, the SVM showed no improvement over the model's internal $p(m|c)$, 
% suggesting the segmentations are already sufficiently mode-specific, 
in line with the theory that modes serve to organise repertoire also in memory \cite[ch.~3]{atkinson2008critical}.
The base NHPYLM model was slightly worse, but did outperform all baselines except for syllables on responsories with absolute pitches (though again barely), further suggesting that despite having more data to estimate a good representation, conditioning on mode would have been a more effective principle for organising repertoire in memory.
NHPYLMClasses also comes very close to the ``upper bound'' overlapping n-gram performance,
suggesting that there may not be much room for improvement on mode classification.

\section{Remembering a Manuscript}
\label{sec:dkaug}

\begin{table}[t]
    \centering
    \setlength{\tabcolsep}{5.5pt}
    \begin{tabular}{lcc|cc}
        \textbf{D-KA Aug. LX data} & \multicolumn{2}{c|}{\textbf{Pitches}} & \multicolumn{2}{c}{\textbf{Intervals}} \\ 
        \cline{2-5}
        \textbf{(Single manuscript)} & \textbf{Ant.} & \textbf{Res.} & \textbf{Ant.} & \textbf{Res.} \\
        \hline
        \textit{Classical approach} & \textit{\textbf{85.5}} & \textit{81.9} & -- & -- \\
        \hline
        4-gram & 82.1 & 77.8 & 70.7 & 64.5 \\
        Syllables & 83.3 & 82.8 & 61.0 & 71.7 \\
        Words & 78.8 & 70.4 & 64.1 & 61.9 \\
        NHPYLM & \textbf{\textit{86.0}} & \textbf{\textit{83.6}} & 73.3 & 72.6 \\  
        NHPYLM-Cl, no SVM & 85.3 & \textbf{84.0} & \textbf{80.2} & \textbf{78.6} \\ 
        NHPYLM-Cl & \textbf{86.1} & \textbf{\textit{83.6}} & \textbf{\textit{79.9}} & \textbf{\textit{78.5}} \\
        \hline
        \textit{Overlapping n-grams, 1-7} & \textit{87.0} & \textit{86.9} & \textit{81.1} & \textit{76.8} \\
    \end{tabular}
    \caption{Mode classification scores for antiphon and responsory melodies from the manuscript D-KA~Aug.~LX, encoded both as sequences of pitches and intervals.}
    \label{tab:dka_aug_f1}
\end{table}

We now consider a computational scenario inspired by a monk needing to memorise melodies for the liturgical environment he belongs to. In our case, this would be in late medieval Zwiefalten, documented by the liturgical book D-KA Aug.~LX. Suppose the monk has already learned 70\% of the melodies (training set): how difficult is it then to memorise the remaining ones?
%These new melodies differ from the training ones, as they come from the same source, eliminating the possibility that the monk might encounter melodies with the same Cantus ID he already saw but from a different source, as seen in Section~\ref{subsec:results}.

\subsection{Mode classification in a single manuscript}

Table~\ref{tab:dka_aug_f1} shows that segmentation based on NHPYLMClasses is again the most accurate model for mode classification and, on responsories with interval representations, outperforms the ``cheating'' overlapping n-grams features, with its internal mode classifier performing slightly better than an SVM on top of the resulting segmentations.

The drop in performance compared to experiments on the full CantusCorpus may have two main reasons: first, it is a much smaller dataset, and second,
only very few Cantus IDs have multiple instances in one liturgical book,
so it eliminates issues with multiple versions of a melody randomly assigned to both the training and test sets. To isolate these influences, we ran two separate experiments:
(1) assigning all instances of a Cantus ID only to one of the train/dev/test sets;
(2) uniformly subsampling CantusCorpus to the size of D-KA Aug.~LX.
In experiment (1), $3 \%$ of performance was lost, and in (2), $4 \%$.
This corresponds surprisingly well to the approximately $7 \%$ drop overall (with pitch representation).\footnote{See \url{github.com/lanzv/chant-modality-with-nhpylms}.}
%With intervals, (1) causes a drop by $5 \%$, (2) by $10 \%$, 
%but the overall drop is only approx. $10 \%$, so these are not as independent.

%for representing (and therefore predicting) an unknown melody’s mode, even outperforming the classical mode classification approach. 
%The difference between NHPYLM and NHPYLMClasses methods is significantly larger in interval-based segmentation. 
%This highlights the importance of considering mode-specific segmentation strategies, especially when working with intervals, rather than relying solely on non-recurring subsequences of intervals across different modes.
%NHPYLMClasses on interval-encoded responsories even surpasses the Overlapping n-grams upper bound estimate.%\footnote{The Words baseline becomes worse than Syllables also for antiphons. It is possible that the preference for word segments in \cite{cornelissen2020mode} was an artifact of different versions of the same melody both in the test and training sets.}

\subsection{Perplexity}
\label{subsec:perplexity}

\begin{table}[t]
    \centering
    \setlength{\tabcolsep}{4pt}
    \begin{tabular}{lcc}
        \hline
        \textbf{Perplexity on D-KA Aug. LX} & \textbf{Antiphons} & \textbf{Resps.} \\
        \hline
        NHPYLM\textit{---intervals} & 20.0 & 17.7 \\
        NHPYLMClasses\textit{---intervals} & 16.1 & 14.3  \\
        NHPYLM\textit{---pitches} & 15.4 & 13.5 \\
        NHPYLMClasses\textit{---pitches} & \textbf{11.8} & \textbf{9.9} \\
        \hline
    \end{tabular}
    \caption{Perplexities for PY-based segmentations on pitch and interval melody encodings on D-KA Aug. LX.}
    \label{tab:perplexity_scores}
\end{table}

We have so far been measuring segmentation quality indirectly, via mode classification.
However, we can also measure its memorisation efficiency directly through \textit{perplexity}, a common metric in language modelling \cite{jelinek1990self} that follows directly from Shannon's coding theorem, and so is a measure of compression (lower is better). Intuitively, it reflects the model’s average uncertainty - the number of equally likely choices it has at each step. It is defined as:
\begin{equation*}
    \text{perplexity}(s_1\dots s_N) = 2^{-\frac{1}{N} \sum_{i=1}^N \log_2 p(s_i \mid h_{i})},
\end{equation*}
where $s_i$ is the $i$-th segment of the given segmentation and $h_i$ is its context.

%In addition to the ability to accurately predict modes, it is crucial that the creation of segmentation of new melodies based on previous segmentations remains as straightforward and simple for a monk as possible. 
%To assess this, 
Table~\ref{tab:perplexity_scores} compares the perplexity of the models. 
NHPYLMClasses leads to more efficient memorisation of melody
than the standard NHPYLM method.\footnote{Though under MDL one should penalize it for having 8 vocabularies instead of a single shared one.}
Interestingly, antiphons appear to be more challenging for effective segmentation than responsories. This is quite in line with the difficulty in identifying
melodic families in antiphons \cite{gevaert1895melopee,merkley1988families},
and conversely the fact that melodic formulas were first described in responsories
\cite{hiley1993western,frere1901antiphonale,helsen2008great}.

Additionally, pitch representation appears easier to remember compared to interval representation, even though their state space is larger.
This may be because in the interval representation, the same segment can occur at different
positions in the pitch system (f-g-f-f equivalent to d-e-d-d), so the conditional entropy of its continuation cannot decrease as much. It is also in line with performance practice:
singers should have been always aware of which pitch within the system they
were singing, as evidenced by prevalent techniques such as the Guidonian hand \cite[ch.~6]{atkinson2008critical} \cite[p.469]{hiley1993western} and solmisation \cite[pp.467--468]{hiley1993western}.

We find a correlation between perplexity values
in Table~\ref{tab:perplexity_scores} and the corresponding mode classification scores
in Table~\ref{tab:dka_aug_f1} of $-0.77$ (and over $-0.88$ within the genres individually), offering initial empirical support for the hypothesised relationship between modality and efficient memorisation, consistent with historical perspectives \cite{lanz2023}.

\subsection{Modal Identity and Memorization in Segments}

\begin{figure}[t] 
  \begin{subfigure}{\linewidth}
    \includegraphics[alt={Two plots showing how melodic formulaicity and modal identity are distributed across the course of antiphon melodies, from beginning to end. For formulaicity, average segment lengths in which each relative position within an antiphon are shown. For modal identity, the average proportion of segments occurring only within antiphons of the same mode is shown. Formulaicity has a small spike at the beginning of antiphon melodies and a slightly larger one at the end, while modal identity only spikes at the beginning of melodies.},width=\linewidth]{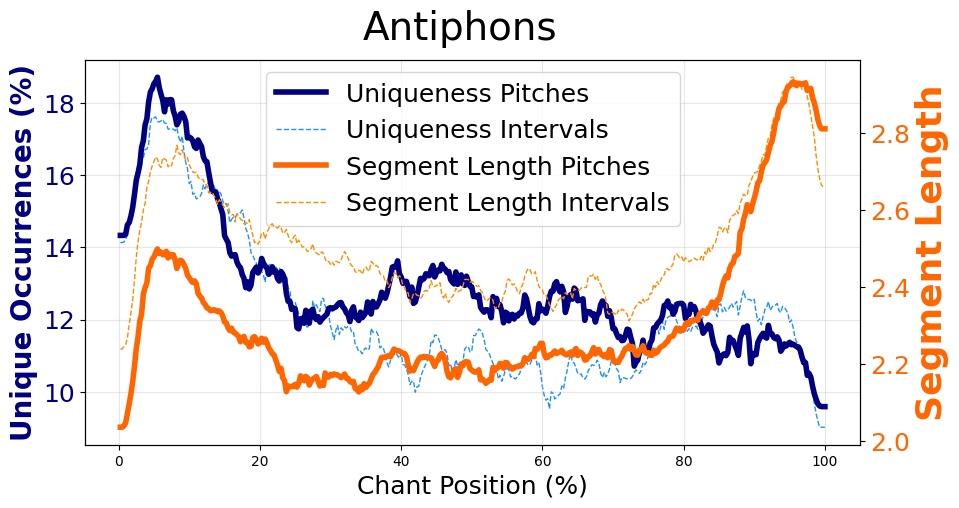}
    %\caption{Average segment lengths (orange) and average segment modal uniqueness (blue) for antiphon data, on both pitches and intervals.} 
    \label{img:antiphons_segments_stats}
  \end{subfigure}
  %\vskip
  \begin{subfigure}{\linewidth}
    \includegraphics[alt={Two plots showing how melodic formulaicity and modal identity are distributed across the course of responsory melodies, from beginning to end. For formulaicity, average segment lengths in which each relative position within a responsory are shown. For modal identity, the average proportion of segments occurring only within responsories of the same mode is shown. Compared to antiphons, formulaicity is very different: there is no spike at the beginning, but a much larger spike towards the end of responsory melodies. However, modal identity is doing nothing of interest.},width=\linewidth]{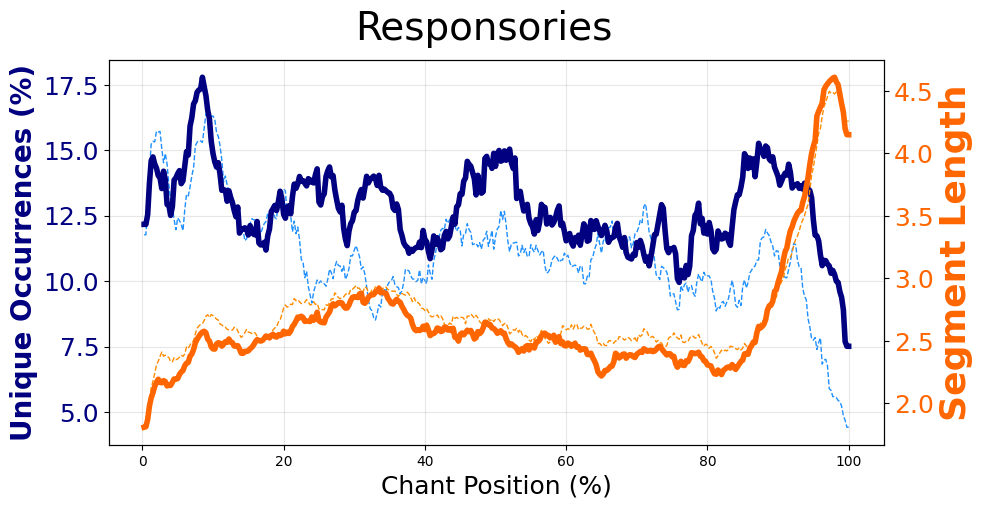} 
    %\caption{Average segment lengths (orange) and average segment modal uniqueness (blue) for responsory data, on both pitches and intervals.} 
    \label{img:responsories_segments_stats}       
  \end{subfigure}
  \caption{Distribution of average segment length (orange; values right) 
           and average segment modal uniqueness (blue; values left)
           across relative positions in melodies. 
           %Larger average lengths correspond
           %to greater formulaicity; higher segment uniqueness corresponds to a stronger
           %relationship between that particular part of the melody and its entire mode
           %(for a given segment to be unique to melodies of the given mode is a strong requirement; this relationship to mode is also encoded in the segment sequence model).
           Note the difference in segment length scales -- responsories are more formulaic. The legend in the antiphon plot applies to both.}
  \label{img:segment_distributions}
\end{figure}

%We turn our attention to how different \textit{parts} of a melody are segmented. 
Is modality, formulaicity, and thus ``centonisability'' more prominent in some parts of the melody than others, and does it relate to mode? This question relates to mode and performance practice. An antiphon was sung before and after a psalm, and psalms were sung on one of just a few simple psalm tones highly specific to individual modes. So, it would have been appropriate for the end (1st antiphon performance) and beginning (2nd antiphon performance) to have a ``compatible'' melody at its interfaces with the psalm tone to which it was assigned (as evidenced, again, by tonaries). For the responsories, instead of a psalm there is a responsory verse, which still has relatively few melodies for each mode but more than psalm tones, and after the verse often only the second half of the responsory (the ``respond'') would be sung.

We plot the average segment length in which a note at a given relative position in the chant melody participates in Figure~\ref{img:segment_distributions} (orange). High average segment length indicate more formulaic melodies. 
For antiphons, we see a small spike in segment length at both ends; for responsories, we see a more prominent spike at the end, corresponding well to the respective performance contexts.

The average uniqueness of the discovered segments to individual modes (Figure~\ref{img:segment_distributions}, in blue) is not as structured.
For the antiphons, we see a spike in uniquess at the beginning, which corresponds to singing the antiphon after the psalm, but at the end of the antiphon (corresponding to the start of the psalm, and highest formulaicity) modal uniqueness drops yet lower. However, psalm tones for some mode pairs have an identical or near-identical start  \cite[p.~59--60]{hiley1993western}, so the performance context makes unique formulas less likely.

\begin{figure}[t]
    \centering
    \includegraphics[width=0.95\linewidth]{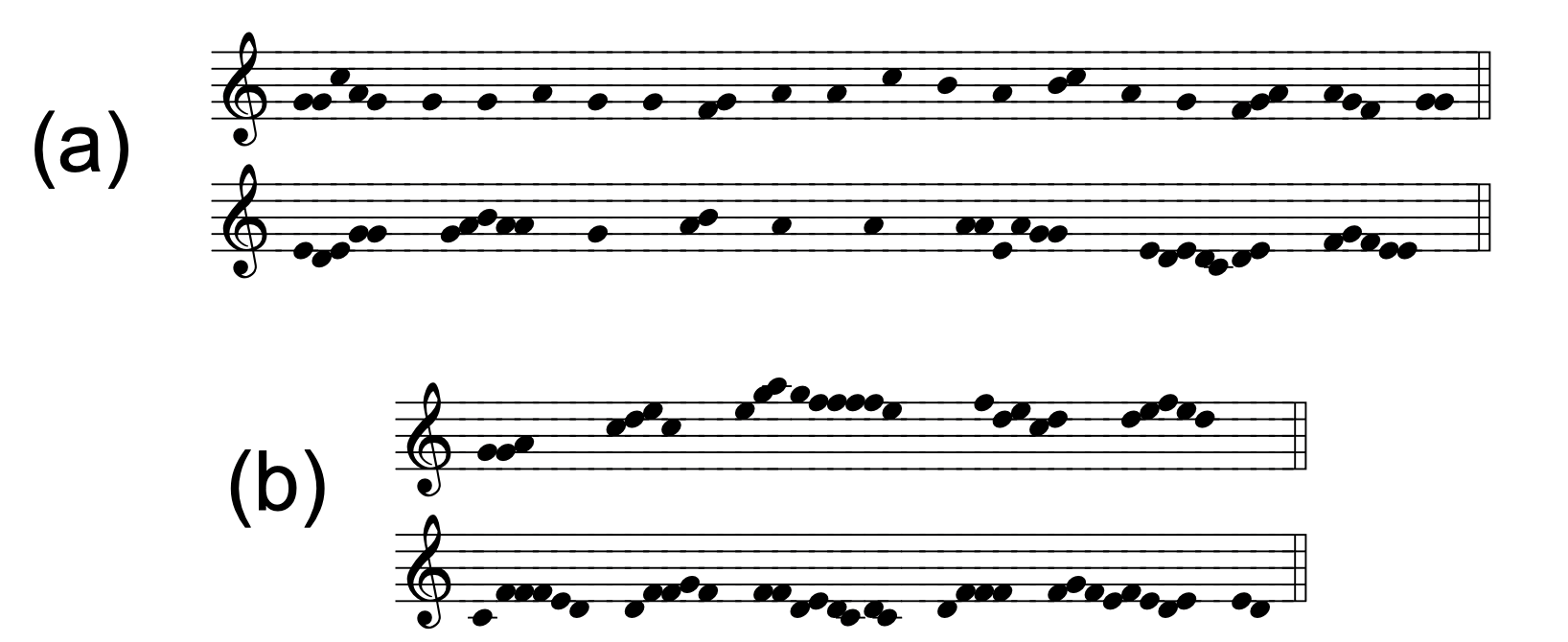}
    \caption{Comparing segmentation patterns: (a) two outputs of our model---a typical case and a rarer, more ``centonised'' one; (b) segmentation proposed by Levy \cite{levy1970italian}.}
    \label{fig:segmentations}
\end{figure}

% ggkhg g g h g g fg h h k j h jk h g fgh hgf gg

% edegg ghjhh g hi h h hhehgg ededcde fgfee

% fg h g fg f f d ff g gf

% \noindent
% Responsories exhibit similar segmentation patterns.
% The second example is truly maximal and only short antiphons in the e-modes exhibit this behaviour. This does not resemble centonisation patterns \cite{levy1970italian,nuttall2019contributing,wellesz1947words}
% at all.

\section{Conclusions}
\label{sec:conclusion}

Building on the memorisation of Gregorian chant, we take an information-theoretic approach and implement NHPYLMs for unsupervised melodic segmentation. The resulting segmentations achieve state-of-the-art mode classification on CantusCorpus v0.2, and in a ``monk simulation'' using D-KA Aug.~LX, also reveal how performance contexts shape formulaicity and modal identity.

%Building on the memorised performance practice of Gregorian chant,
%we take an information-theoretic approach to find optimal segmentations of its melody and
%Taking inspiration from the Bayesian interpretation of the two-part MDL principle, we 
%implement and apply NHPYLMs, unsupervised segmentation language models.
%
%The resulting segmentations achieve state-of-the-art mode classification results,
%both over CantusCorpus v0.2 and in a ``monk simulation'' on a single monastic source, D-KA Aug.~LX.
%We also saw different performance contexts reflected in resulting patterns of formulaicity and modal identity within melodies.

The correlation tentatively found between perplexity, as ``ease of memorisation'', and mode classification F1-score
is initial empirical evidence for the relationship between melody segmentation and modality,
in accordance with the music-historical argument for this link \cite{atkinson2008critical,cornelissen2020mode,lanz2023}. 
Pilot experiments with other melody encodings\footnote{E.g., collapsing repeated notes -- see results in the Supplementaries.} suggest this holds across conditions, warranting further study.
%Pilot experiments with other encodings of melody\footnote{See results in the Supplementaries.} %(e.g. different handling of liquescences and merging repeated notes) 
%show this correlation holds across different conditions, and we are excited to explore this link further.
%
%Responsories, though longer and usually described as more complex melodies than antiphons, proved simpler to remember than antiphons. %\textbf{TODO (...) average perplexity per position?}
%

% we segmented Gregorian chant melodies to improve modality description. To achieve this, we applied the nonparametric Bayesian method NHPYLM, based on the minimum description length principle. Additionally, we introduced an extension, NHPYLMClasses, which incorporates the concept of modality during segmentation and enables mode prediction for a given chant melody.

% We generated new best segmentation in terms of mode classification. We also demonstrated that while pitch representation remains slightly superior for describing modality, interval representation can still be meaningful. Furthermore, we examined segmentation behaviour within a single chant source, exploring how it may aid monks in memorization.

%\textbf{TODO:Future work.}

%The Dirichlet Process, with exponential expected no. of classes instead of power-law, might be even better than PY. Given the complexity of the model implementation, however, we focused on PY and its nested hierarchical version.

%The correlation between mode classification scores and perplexity is a new hint of empirical evidence to the discussion on what is Gregorian mode. 

Memory efficiency of baseline segmentations could be approximated 
with a bigram language model with Interpolated Kneser-Ney smoothing \cite{ney1994smoothing} trained on the optimal segmentations.
KN smoothing is closely related to the PY process \cite{teh2006}.
Pilot results showed its perplexity differed by no more than 0.5 from NHPYLM's on D-KA Aug.~LX.
%and pilot experiments showed that, for NHPYLM segmentations, 
%the perplexity estimated this way is not further than 0.5 from perplexities computed on D-KA Aug. LX from the NHPYLM itself.

%NHPYLM models can be used to obtain memory-based segmentations
%of other repertoires: to verify that it can discover centonisation where we know from musicology
%that it should be present \cite{nuttall2019contributing,wellesz1947words,huglo1985beneventan}
%and to make an information-theoretic comparison between Gregorian repertoire and other traditions; however, for this, datasets first need to be assembled.
NHPYLMs can provide memory-based segmentations of other repertoires -- both to detect centonisation where it is musicologically known to exist \cite{nuttall2019contributing,wellesz1947words,huglo1985beneventan}, and to enable information-theoretic comparisons with Gregorian chant. However, suitable datasets are needed first.

%Having provided the statistical context, 
%we finally show examples of the resulting segmentations.
%Do the discovered optimal segmentations correspond to proposed centonisation patterns? Typically, no. 

%\textbf{TODO: Move these sentences to conclusions?}
However, despite all the reasons to believe Gregorian chant could be centonised, and despite the empirical indications of a close relationship between segmentation, memory, and modality, 
the discovered optimal segments are far from a convincing centonisation of chant.
Comparing a typical model output, and an output with maximum formulaicity, with centonisation proposed in chant scholarship in Figure~\ref{fig:segmentations}, we see that the discovered segmentation patterns are far more fragmentary than what has been proposed for Gregorian chant \cite{levy1970italian} and what is known from Byzantine formulae \cite{wellesz1947words}.\footnote{Full segmentation results on D-KA Aug. LX are available in Supplementary materials.}
This opens the intriguing possibility that institutional rules and practices were able
to significantly counteract processes inherent in oral transmission,
making the evolutionary processes of Gregorian chant different
from other musical traditions.
The mystery of constructing Gregorian melodies remains unresolved.

\section{Acknowledgments}

The authors are supported by the Social Sciences and Humanities Research Council of Canada by the grant no. 895-2023-1002, Digital Analysis of Chant Transmission, and by the SVV project number 260 698. The second author additionally acknowledges the support by the project ``Human-centred AI for a Sustainable and Adaptive Society'' (reg. no.: CZ.02.01.01/00/23\_025/0008691), co-funded by the European Union. The computing infrastructure is provided by the LINDAT/CLARIAH-CZ Research Infrastructure (https://lindat.cz), supported by the Ministry of Education, Youth and Sports of the Czech Republic (Project No. LM2023062).

\section{Ethics Statement}

This study did not involve human participants, personal data, or sensitive information. All data  consists of transcriptions of publicly available historical sources. No concerns related to privacy, consent, or bias apply. % in this research.
We, however, note that we aim to contribute to the understanding of chant without reinforcing any cultural or religious biases; our work in no way implies that Gregorian chant having different transmission patterns would be a value judgment. 
%The proposed methods and findings are intended solely for academic and musicological purposes.

\section{Data and code accessibility}
\label{sec:data_access}

The CantusCorpus v0.2 dataset \cite{cornelissen2020studying} that we used is available at: \url{https://github.com/bacor/cantuscorpus/releases/tag/v0.2}
The NHPYLM model code is available at: \url{https://github.com/lanzv/nhpylm}. 
The experiment code, including more detailed results, is available at: \url{https://github.com/lanzv/chant-modality-with-nhpylms}.

% \section{Camera Ready Preparation}\label{sec:cameraready}

% \textbf{The camera-ready version should include the names, affiliations and email addresses of the authors.} Authors' names are centered. The lead author's name is to be listed first (left-most), and the co-authors' names after. If the addresses for all authors are the same, include the address only once, centered. If the authors have different addresses, put the addresses, evenly spaced, under each authors' name. To display the author information in \LaTeX, please remove the global option `submission' when importing the `ismir' package (i.e., \texttt{\textbackslash usepackage\{ismir\}} in line 16).

% \textbf{Please make sure that the author names, paper title and proceedings title are shown correctly in the copyright notice.} For \LaTeX\ users, the proceedings title will be automatically loaded when you remove the global option `submission' when importing the `ismir' package (i.e., \texttt{\textbackslash usepackage\{ismir\}} in line 16). For Word users, you will need to manually replace ``submitted to \textit{ISMIR}, 2025'' to ``in \textit{Proc. of the 26th Int. Society for Music Information Retrieval Conf.}, Daejeon, South Korea, 2025.''

% \textbf{You must also remove all line numbers from the final camera-ready version.} This can be done in \LaTeX\ by removing the global option `submission' when importing the ismir package (i.e., \texttt{\textbackslash usepackage\{ismir\}} in line 16). This can be done in Microsoft Word by selecting ``Layout > Line Numbers > None.''

% For BibTeX users:
\bibliography{bibliography}

\begin{thebibliography}{10}
\providecommand{\url}[1]{#1}
\csname url@samestyle\endcsname
\providecommand{\newblock}{\relax}
\providecommand{\bibinfo}[2]{#2}
\providecommand{\BIBentrySTDinterwordspacing}{\spaceskip=0pt\relax}
\providecommand{\BIBentryALTinterwordstretchfactor}{4}
\providecommand{\BIBentryALTinterwordspacing}{\spaceskip=\fontdimen2\font plus
\BIBentryALTinterwordstretchfactor\fontdimen3\font minus \fontdimen4\font\relax}
\providecommand{\BIBforeignlanguage}[2]{{%
\expandafter\ifx\csname l@#1\endcsname\relax
\typeout{** WARNING: IEEEtran.bst: No hyphenation pattern has been}%
\typeout{** loaded for the language `#1'. Using the pattern for}%
\typeout{** the default language instead.}%
\else
\language=\csname l@#1\endcsname
\fi
#2}}
\providecommand{\BIBdecl}{\relax}
\BIBdecl

\bibitem{hiley1993western}
D.~Hiley, \emph{Western plainchant: a handbook}.\hskip 1em plus 0.5em minus 0.4em\relax Oxford, United Kingdom: Clarendon Press, 1993.

\bibitem{hallas2021offices}
R.~Hallas, \emph{Two rhymed offices composed for the feast of the Visitation of the Blessed Virgin Mary: comparative study and critical edition}.\hskip 1em plus 0.5em minus 0.4em\relax Bangor University (United Kingdom), 2021.

\bibitem{merkley1988families}
\BIBentryALTinterwordspacing
P.~Merkley, ``Tonaries and melodic families of antiphons,'' \emph{Journal of the Plainsong and Mediaeval Music Society}, vol.~11, p. 13–24, Jan. 1988. [Online]. Available: \url{http://dx.doi.org/10.1017/S0143491800001136}
\BIBentrySTDinterwordspacing

\bibitem{atkinson2008critical}
C.~M. Atkinson, \emph{The critical nexus: tone-system, mode, and notation in early medieval music}.\hskip 1em plus 0.5em minus 0.4em\relax Oxford University Press, 2008.

\bibitem{glasenapp2020pray}
J.~Glasenapp, \emph{To Pray without Ceasing: A Diachronic History of Cistercian Chant in the Beaupr{\'e} Antiphoner (Baltimore, Walters Art Museum, W. 759--762)}.\hskip 1em plus 0.5em minus 0.4em\relax Columbia University, 2020.

\bibitem{huglo1971tonaires}
M.~Huglo, ``Les tonaires,'' \emph{Inventaire, Analyse, Comparaison. Paris: Soci{\'e}t{\'e} fran{\c{c}}aise de musicology}, pp. 132--40, 1971.

\bibitem{cornelissen2020mode}
B.~Cornelissen, W.~H. Zuidema, and J.~A. Burgoyne, ``Mode classification and natural units in plainchant.'' in \emph{Proceedings of the 21st Int. Society for Music Information Retrieval Conf.}, Montreal, Canada, 2020, pp. 869--875.

\bibitem{mori2001conflicting}
H.~Mori, \emph{Conflicting modal assignments of office antiphons: A comparative study of seven Germanic sources.}\hskip 1em plus 0.5em minus 0.4em\relax National Library of Canada= Bibliotheque nationale du Canada, Ottawa, 2001.

\bibitem{helsen2009use}
K.~Helsen, ``The use of melodic formulas in responsories: constancy and variability in the manuscript tradition,'' \emph{Plainsong \& Medieval Music}, vol.~18, no.~1, pp. 61--76, 2009.

\bibitem{karp1998aspects}
T.~Karp, \emph{Aspects of orality and formularity in Gregorian chant}.\hskip 1em plus 0.5em minus 0.4em\relax Northwestern University Press, 1998.

\bibitem{Helsen2023}
\BIBentryALTinterwordspacing
K.~Helsen, M.~Daley, and J.~Schindler, ``The sticky riff: Quantifying the melodic identities of medieval modes,'' \emph{Empirical Musicology Review}, vol.~16, no.~2, p. 312–325, Mar. 2023. [Online]. Available: \url{http://dx.doi.org/10.18061/emr.v16i2.7357}
\BIBentrySTDinterwordspacing

\bibitem{gevaert1895melopee}
F.-A. Gevaert, \emph{La Mélopée Antique dans le Chant de l'Église Latine}.\hskip 1em plus 0.5em minus 0.4em\relax Ad. Hoste, 1895.

\bibitem{lanz2023}
V.~Lanz and J.~Haji{\v{c}}, ``Text boundaries do not provide a better segmentation of gregorian antiphons,'' in \emph{Proceedings of the 10th International Conference on Digital Libraries for Musicology}, 2023, pp. 72--76.

\bibitem{lacoste2022cantus}
\BIBentryALTinterwordspacing
D.~Lacoste, ``{The Cantus Database and Cantus Index Network},'' in \emph{{The Oxford Handbook of Music and Corpus Studies}}.\hskip 1em plus 0.5em minus 0.4em\relax Oxford University Press, 2022. [Online]. Available: \url{https://doi.org/10.1093/oxfordhb/9780190945442.013.18}
\BIBentrySTDinterwordspacing

\bibitem{cornelissen2020studying}
B.~Cornelissen, W.~Zuidema, and J.~A. Burgoyne, ``Studying large plainchant corpora using chant21,'' in \emph{7th International Conference on Digital Libraries for Musicology}, 2020, pp. 40--44.

\bibitem{ferretti1934estetica}
\BIBentryALTinterwordspacing
P.~Ferretti, \emph{Estetica gregoriana: ossia, Trattato delle forme musicali del canto gregoriano}, ser. Estetica gregoriana: ossia, Trattato delle forme musicali del canto gregoriano.\hskip 1em plus 0.5em minus 0.4em\relax Pontificio istituto di musica sacra, 1934, no. sv. 1. [Online]. Available: \url{https://books.google.cz/books?id=vOWCnQEACAAJ}
\BIBentrySTDinterwordspacing

\bibitem{levy1970italian}
K.~Levy, ``The italian neophytes' chants,'' \emph{Journal of the American Musicological Society}, vol.~23, no.~2, pp. 181--227, 1970.

\bibitem{treitler1975centonate}
L.~Treitler, ``Centonate" chant:" {\"u}bles flickwerk" or" e pluribus unus?'' \emph{Journal of the American Musicological Society}, vol.~28, no.~1, pp. 1--23, 1975.

\bibitem{hucke1980toward}
H.~Hucke, ``Toward a new historical view of gregorian chant,'' \emph{Journal of the American Musicological Society}, vol.~33, no.~3, pp. 437--467, 1980.

\bibitem{shanahan2019examining}
D.~Shanahan and J.~Albrecht, ``Examining the effect of oral transmission on folksongs,'' \emph{Music Perception: An Interdisciplinary Journal}, vol.~36, no.~3, pp. 273--288, 2019.

\bibitem{anglada2022studying}
M.~Anglada-Tort, P.~M. Harrison, and N.~Jacoby, ``Studying the effect of oral transmission on melodic structure using online iterated singing experiments,'' \emph{bioRxiv}, pp. 2022--05, 2022.

\bibitem{popescu2024evolution}
\BIBentryALTinterwordspacing
T.~Popescu and M.~Rohrmeier, ``Core principles of melodic organisation emerge from transmission chains with random melodies,'' \emph{Evolution and Human Behavior}, vol.~45, no.~6, p. 106619, 2024. [Online]. Available: \url{https://www.sciencedirect.com/science/article/pii/S1090513824000953}
\BIBentrySTDinterwordspacing

\bibitem{froger1978critical}
D.~J. Froger, ``The critical edition of the roman gradual by the monks of solesmes,'' \emph{Journal of the Plainsong \& Mediaeval Music Society}, vol.~1, pp. 81--97, 1978.

\bibitem{hajic2023}
\BIBentryALTinterwordspacing
J.~Hajič~jr., G.~A. Ballen, K.~H. Mühlová, and H.~Vlhová-Wörner, ``{Towards Building a Phylogeny of Gregorian Chant Melodies},'' in \emph{{Proceedings of the 24th International Society for Music Information Retrieval Conference}}.\hskip 1em plus 0.5em minus 0.4em\relax ISMIR, Dec. 2023, pp. 571--578. [Online]. Available: \url{https://doi.org/10.5281/zenodo.10340442}
\BIBentrySTDinterwordspacing

\bibitem{nuttall2019contributing}
T.~Nuttall, M.~Garc{\'\i}a~Casado, V.~N{\'u}{\~n}ez~Tarifa, R.~Caro~Repetto, and X.~Serra, ``Contributing to new musicological theories with computational methods: the case of centonization in {A}rab-{A}ndalusian music,'' in \emph{20th Conference of the International Society for Music Information Retrieval (ISMIR 2019): 2019 Nov 4-8; Delft, The Netherlands.[Canada]: ISMIR; 2019. p. 223-8.}\hskip 1em plus 0.5em minus 0.4em\relax International Society for Music Information Retrieval (ISMIR), 2019.

\bibitem{wellesz1947words}
E.~Wellesz, ``Words and music in byzantine liturgy,'' \emph{The Musical Quarterly}, vol.~33, no.~3, pp. 297--310, 1947.

\bibitem{connolly1972introits}
T.~H. Connolly, ``Introits and archetypes: Some archaisms of the old roman chant,'' \emph{Journal of the American Musicological Society}, vol.~25, no.~2, pp. 157--174, 1972.

\bibitem{dyer1998tropis}
J.~Dyer, ``Tropis semper variantibus: Compositional strategies in the offertories of old roman chant,'' \emph{Early Music History}, vol.~17, pp. 1--60, 1998.

\bibitem{huglo1985beneventan}
M.~Huglo, ``The old beneventan chant,'' \emph{Studia Musicologica Academiae Scientiarum Hungaricae}, vol.~27, no. Fasc. 1/4, pp. 83--95, 1985.

\bibitem{chakraborty2021hindustani}
S.~Chakraborty, S.~Tewari, A.~Rahman, M.~Jamal, A.~Lipi, A.~Chakraborty, A.~Nanda, and P.~Shukla, \emph{Hindustani classical music: a historical and computational study}.\hskip 1em plus 0.5em minus 0.4em\relax Sanctum Books, 2021.

\bibitem{rissanen1998stochastic}
J.~Rissanen, \emph{Stochastic complexity in statistical inquiry}.\hskip 1em plus 0.5em minus 0.4em\relax World scientific, 1998, vol.~15.

\bibitem{grunwald2007minimum}
P.~D. Gr{\"u}nwald, \emph{The minimum description length principle}.\hskip 1em plus 0.5em minus 0.4em\relax MIT press, 2007.

\bibitem{barron1991minimum}
A.~R. Barron and T.~M. Cover, ``Minimum complexity density estimation,'' \emph{IEEE transactions on information theory}, vol.~37, no.~4, pp. 1034--1054, 1991.

\bibitem{Goldwater2009Bayesian}
\BIBentryALTinterwordspacing
S.~Goldwater, T.~L. Griffiths, and M.~Johnson, ``A bayesian framework for word segmentation: Exploring the effects of context,'' \emph{Cognition}, vol. 112, no.~1, pp. 21--54, 2009. [Online]. Available: \url{https://www.sciencedirect.com/science/article/pii/S0010027709000675}
\BIBentrySTDinterwordspacing

\bibitem{teh2006}
Y.~W. Teh, ``A bayesian interpretation of interpolated kneser-ney,'' School of Computing, National University of Singapore, Technical Report TRA2/06, 2006.

\bibitem{walter2013unsupervised}
O.~Walter, R.~Haeb-Umbach, S.~Chaudhuri, and B.~Raj, ``Unsupervised word discovery from phonetic input using nested pitman-yor language modeling,'' in \emph{ICRA Workshop on Autonomous Learning}, 2013.

\bibitem{takeda2018word}
R.~Takeda, K.~Komatani, and A.~I. Rudnicky, ``Word segmentation from phoneme sequences based on pitman-yor semi-markov model exploiting subword information,'' in \emph{2018 IEEE Spoken Language Technology Workshop (SLT)}.\hskip 1em plus 0.5em minus 0.4em\relax IEEE, 2018, pp. 763--770.

\bibitem{araki2012online}
T.~Araki, T.~Nakamura, T.~Nagai, S.~Nagasaka, T.~Taniguchi, and N.~Iwahashi, ``Online learning of concepts and words using multimodal lda and hierarchical pitman-yor language model,'' in \emph{2012 IEEE/RSJ International Conference on Intelligent Robots and Systems}.\hskip 1em plus 0.5em minus 0.4em\relax IEEE, 2012, pp. 1623--1630.

\bibitem{mochihashi2009}
\BIBentryALTinterwordspacing
D.~Mochihashi, T.~Yamada, and N.~Ueda, ``{B}ayesian unsupervised word segmentation with nested {P}itman-{Y}or language modeling,'' in \emph{Proceedings of the Joint Conference of the 47th Annual Meeting of the {ACL} and the 4th International Joint Conference on Natural Language Processing of the {AFNLP}}.\hskip 1em plus 0.5em minus 0.4em\relax Suntec, Singapore: Association for Computational Linguistics, Aug. 2009, pp. 100--108. [Online]. Available: \url{https://aclanthology.org/P09-1012}
\BIBentrySTDinterwordspacing

\bibitem{sawada2020unsupervised}
\BIBentryALTinterwordspacing
S.~Sawada, K.~Yoshii, and K.~Hirata, ``Unsupervised melody segmentation based on a nested {P}itman-{Y}or language model,'' in \emph{Proceedings of the 1st Workshop on NLP for Music and Audio (NLP4MusA)}, S.~Oramas, L.~Espinosa-Anke, E.~Epure, R.~Jones, M.~Sordo, M.~Quadrana, and K.~Watanabe, Eds.\hskip 1em plus 0.5em minus 0.4em\relax Online: Association for Computational Linguistics, 16 Oct. 2020, pp. 59--63. [Online]. Available: \url{https://aclanthology.org/2020.nlp4musa-1.12/}
\BIBentrySTDinterwordspacing

\bibitem{PitmanYor1997}
\BIBentryALTinterwordspacing
J.~Pitman and M.~Yor, ``The two-parameter poisson-dirichlet distribution derived from a stable subordinator,'' \emph{The Annals of Probability}, vol.~25, no.~2, Apr. 1997. [Online]. Available: \url{http://dx.doi.org/10.1214/aop/1024404422}
\BIBentrySTDinterwordspacing

\bibitem{Hallam2017musicPsychology}
S.~Hallam, I.~Cross, and M.~Thaut, Eds., \emph{\BIBforeignlanguage{en}{The oxford handbook of music psychology}}, 2nd~ed., ser. Oxford Library of Psychology.\hskip 1em plus 0.5em minus 0.4em\relax London, England: Oxford University Press, Dec. 2017.

\bibitem{jelinek1990self}
F.~Jelinek, ``Self-organized language modeling for speech recognition,'' \emph{Readings in speech recognition}, pp. 450--506, 1990.

\bibitem{frere1901antiphonale}
W.~H. Frere, \emph{Antiphonale Sarisburiense: a reproduction in facsimile of a manuscript of the 13th century, with a dissertation and analytical index}.\hskip 1em plus 0.5em minus 0.4em\relax Gregg Press Limited, 1901.

\bibitem{helsen2008great}
K.~E. Helsen, ``The great responsories of the divine office: aspects of structure and transmission,'' Ph.D. dissertation, 2008.

\bibitem{ney1994smoothing}
\BIBentryALTinterwordspacing
H.~Ney, U.~Essen, and R.~Kneser, ``On structuring probabilistic dependences in stochastic language modelling,'' \emph{Computer Speech \& Language}, vol.~8, no.~1, p. 1–38, Jan. 1994. [Online]. Available: \url{http://dx.doi.org/10.1006/csla.1994.1001}
\BIBentrySTDinterwordspacing

\end{thebibliography}

\end{document}